# Few-shot fault diagnosis based on multi-scale graph convolution filtering for industry

Mengjie Gan[1], Penglong Lian[2], Zhiheng Su[2], Jiyang Zhang[2], Jialong Huang[1], Benhao Wang[1], Jianxiao Zou[1,2] and Shicai Fan[1,2*]

*Abstract*— Industrial equipment fault diagnosis often encounter challenges such as the scarcity of fault data, complex operating conditions, and varied types of failures. Signal analysis, data statistical learning, and conventional deep learning techniques face constraints under these conditions due to their substantial data requirements and the necessity for transfer learning to accommodate new failure modes. To effectively leverage information and extract the intrinsic characteristics of faults across different domains under limited sample conditions, this paper introduces a fault diagnosis approach employing Multi-Scale Graph Convolution Filtering (MSGCF). MSGCF enhances the traditional Graph Neural Network (GNN) framework by integrating both local and global information fusion modules within the graph convolution filter block. This advancement effectively mitigates the over-smoothing issue associated with excessive layering of graph convolutional layers while preserving a broad receptive field. It also reduces the risk of overfitting in few-shot diagnosis, thereby augmenting the model's representational capacity. Experiments on the University of Paderborn bearing dataset (PU) demonstrate that the MSGCF method proposed herein surpasses alternative approaches in accuracy, thereby offering valuable insights for industrial fault diagnosis in few-shot learning scenarios.

## I. INTRODUCTION

As modern industry progresses, ranging from basic components like bearings, gears, and rotating shafts to their integration into large-scale machinery such as wind turbines, aerospace engines, and high-speed trains, industrial equipment failures are predominantly attributed to internal mechanical or electrical structural defects. These failures can lead to severe casualties and significant economic losses [1]-[2]. The shift towards integrated industrial apparatus and unmanned operations has rendered manual real-time monitoring economically impractical. Consequently, there is an escalating demand for intelligent fault diagnosis to ensure the system's operational health. However, intelligent diagnostic efforts encounter formidable challenges in practical settings: (1) Data Scarcity: In real industrial systems, faults are infrequent, making the artificial generation of fault data prohibitively expensive and disruptive to production; (2) Variability in Working Conditions: Actual data may originate from diverse operating conditions (such as variations in rotation speed and workload) and different machinery (such as various bearings), leading to a wide disparity in the distribution of fault data under different conditions; (3) Complex Fault Phenomena: For instance, bearing faults can exhibit varying degrees of damage, types of damage, and loss criteria. Accurate identification of fault phenomena in real industries is crucial. Therefore, addressing these daunting conditions for fault diagnosis has emerged as a prominent topic and challenge in current research endeavors.

Intelligent fault diagnosis has evolved from traditional machine learning methods like DPCA-SVM [3], OCMLFEM [4], and KNN [5] to advanced deep learning techniques including CNN [6] and LSTM [7], enabling automatic feature extraction for fault identification. Despite their advances, these approaches face challenges in real-world applications due to data scarcity, changing conditions, and complex fault dynamics. Generative Adversarial Network (GAN) [8] offer a solution by generating data when scarce, but their effectiveness depends on the original data quality and computational intensity. Meta-learning, excelling in few-shot learning, follows optimization-based and metric-based paths. Model-Agnostic Meta-Learning (MAML) [9], an optimization-based strategy, optimizes for global parameters in meta-tasks, enhancing convergence and reducing overfitting as demonstrated by Zhang et al. [9]. Jiang et al. [10] introduced a two-branch approach, merging time and frequency data within a prototype network (TBPN) for effective fault diagnosis. Li et al. [11] proposed a WDCNN-based Siamese network for few-shot learning, optimizing through variable targets and depth-separable convolutions, highlighting the method's adaptability and efficiency in complex diagnostic scenarios.

In few-shot learning scenarios with limited samples across categories, traditional metric models, focusing solely on pairwise sample relationships, fall short in harnessing the full potential of known sample connections. Graph Neural Networks (GNNs) address this by integrating metric learning principles to optimize the informational relationships among samples. Within GNNs, both support and query samples are modeled as graph nodes, interconnected through an adjacency matrix to facilitate efficient message exchange between nodes. This approach significantly enhances the utilization of sample relationships, making GNN-based methods [12]-[14] increasingly popular for their effectiveness in small sample settings and their application in fault diagnosis [15]-[16].

The Graph Convolution Filter Block (GCN) forms the foundational structure of graph neural network architectures. To tackle the issue of over-smoothing, which arises from the

This work was partially supported by the Shenzhen Science and Technology Program (no. JCYJ20210324140407021).

[1]Mengjie Gan, Jialong Huang and Benghao Wang are with Shenzhen Institute of Advanced Studies, University of Electronic Science and Technology of China, Shenzhen, Guangdong, China 518000 (e-mail: 202222280530@std.uestc.edu.cn;).

[2]Penglong Lian, Jiyang Zhang and Zhiheng Su are with School of Automation Engineering, University of Electronic Science and Technology of China, Chengdu, Sichuan, China

[1]Jianxiao Zou and Shicai Fan are with Shenzhen Institute for Advanced Study and School of Automation Engineering, University of Electronic Science and Technology of China (UESTC), China (e-mail: jxzou@uestc.edu.cn; shicaifan@uestc.edu.cn).

*Corresponding author.

excessive layering of graph convolution filter blocks, this research introduces a novel fault diagnosis approach based on Multi-Scale Graph Convolution Filtering (MSGCF). Initially, the method employs a convolutional neural network for feature extraction and dimensionality reduction on the original signal, effectively reducing complexity for further modeling. During the rapid stacking phase of graph convolution filtering, it deliberately shares the input information from the previous layer's filter block on a local level, while paralleling the output information of the single-layer filter on a global scale. This strategy is designed to mitigate the accuracy decline associated with over-smoothing and to maintain an adequate receptive field for the nodes, ensuring the comprehensive utilization of sample information. The method in this paper has the following advantages:

**(1) This paper proposes a filtering structure based on multi-scale graph convolution, which balances the contradiction between stacked graph filtering to quickly increase the receptive field and the over-smoothing phenomenon. In the ablation experiment, both the local channel and the global channel are better than the original GNN had a positive effect on the results.**

**(2) The proposed model makes full use of sample information to support message passing between samples compared to traditional measurement models, and has achieved excellent fault diagnosis results in experiments.**

The remainder of this article is structured as follows: Section II introduces fundamental concepts of graph theory and few-shot learning theory. Section III outlines the comprehensive framework of the proposed method. Section IV presents detailed experimental results. Finally, Section V concludes the article.

## II. PRELIMINARIES

### A. Graph definition

The core of graph convolutional networks lies in graph signal processing (GSP), which is essentially the application of discrete signal processing to graph signals[17]. A graph consists of numerous nodes and connecting edges, often used to delineate specific relationships among various entities. In the context of an undirected graph, it is denoted as $G = \{V, E\}$, where $V$ is the set of graph nodes, with $V_i$ denoting the i-th node, and $E$ is the set of edges, with $e_{ij}$ representing the connection between nodes i and j. N indicates the total number of nodes in the graph. The adjacency matrix $A \in R^{N \times N}$ encapsulates the connectivity of the graph, with the element $A_{ij}$ specifying the link between nodes i and j. In an unweighted graph, $A_{ij} = 1$ (if $e_{ij} \in E$); for a weighted graph, $A_{ij}$ denotes the weight associated with the edge $e_{ij}$. Furthermore, the degree matrix $D \in R^{N \times N}$, a diagonal matrix, represents the number of connections to each node, where $D_{ii}$ equates to the sum of the weights of edges connected to node i, $D$ is a real symmetric matrix that satisfies $D_{ij} = 0$, $i \neq j$. The calculation formula of $D_{ii}$ is as follows:

$$D_{ii} = \sum_j A_{ij} \qquad (1)$$

The foundation of graph frequency domain analysis is the Laplacian matrix of the graph, defined as follows:

$$L = D - A \qquad (2)$$

Additionally, there exists a symmetric normalized Laplacian matrix, which is defined as:

$$L_{sym} = D^{-\frac{1}{2}} L D^{\frac{1}{2}} \qquad (3)$$

### B. Graph Fourier Transform

A graph signal represents a mapping from a node set $V$ to an N-dimensional real number domain: V → R. In multi-channel graph signals, as per academic conventions, N denotes the number of graph nodes, and $f$ is the count of signal channels per node, with each graph signal matrix row $X \in R^{N \times f}$ symbolizing the signal on a node. Thus, the graph signal $X$ can be defined as:

$$X = [x_1, x_2, ..., x_N] \qquad (4)$$

where $x_i$ represents the signal of the i-th node in the graph.

Given the Laplacian matrix $L$ is a real symmetric positive semidefinite matrix, it possesses a set of orthogonal eigenvectors $U = \{v_i\}_{i=1}^N$ with corresponding eigenvalues $\{\lambda_i\}_i^N$ being non-negative. By orthogonal diagonalizing of L, we can get:

$$L = U \Lambda U^T = [v_1 \ v_2 \ \cdots \ v_N] \begin{bmatrix} \lambda_1 & & & \\ & \lambda_2 & & \\ & & \ddots & \\ & & & \lambda_N \end{bmatrix} \begin{bmatrix} v_1^T \\ v_2^T \\ \vdots \\ v_N^T \end{bmatrix} \qquad (5)$$

As outlined in the literature [19], the graph Fourier transform (GFT) projects a graph signal onto each eigenvector, which serves as a Fourier basis, to derive a spectrum of Fourier coefficients. This projection effectively constitutes the GFT, and the process for computing the complete set of Fourier coefficients is as follows:

$$\tilde{X} = U^T X = \begin{bmatrix} v_1^T \\ v_2^T \\ \vdots \\ v_N^T \end{bmatrix} X \qquad (6)$$

Since the eigenvector $U$ of the Laplacian matrix is a complete orthogonal basis in a set of N-dimensional space, multiplying the above equation by $U$ on the left, the inverse graph Fourier transform (IGFT) is:

$$X = U \tilde{X} = \sum_{i=1}^N \tilde{x}_i v_i \qquad (7)$$

### C. Graph convolution filtering (GCN)

In graph signal processing, the operation of enhancing or attenuating the intensity of each frequency component in the spectrum of a graph signal is defined as a graph filter. Suppose the graph filter is $H \in R^{N \times N}$, and the output graph signal after graph filtering is $Y \in R^{N \times f}$, then $H$ is equivalent to performing a signal transformation on the input signal X:

$$Y = HX = \sum_{i=1}^{N} (h(\lambda_i)\tilde{x}_i)v_i \qquad (8)$$

The function of the $h(\lambda)$ term is to enhance or attenuate the control signal. Simplifying the above formula, we can get:

$$H = U \begin{bmatrix} h(\lambda_1) & & & \\ & h(\lambda_2) & & \\ & & \ddots & \\ & & & h(\lambda_N) \end{bmatrix} U^T = U\Lambda_h U^T \qquad (9)$$

where $\Lambda_h$ is the frequency response matrix of the graph filter $H$. Compared with the Laplacian matrix $L$, $H$ only changes the value of the diagonal matrix $\Lambda$. Therefore, the graph filter is actually a function that acts on the eigenvalues of the Laplacian matrix, which uses a frequency response function $h(\lambda)$ to adjust the intensity of components at different frequencies (different eigenvalues), different frequency response functions can achieve different filtering effects. In order to achieve different filtering effects, through approximation theory, Taylor expansion-polynomial approximation function can approximate any function, and use a polynomial function of a Laplacian matrix to approximate any filter:

$$H = \sum_{k=0}^{K} h_k L^k$$

$$= U \begin{bmatrix} \sum_{k=0}^{K} h_k \lambda_1^k & & & \\ & \sum_{k=0}^{K} h_k \lambda_2^k & & \\ & & \ddots & \\ & & & \sum_{k=0}^{K} h_k \lambda_N^k \end{bmatrix} U^T \qquad (10)$$

where $h = [h_1, h_2, ..., h_K]$ is the coefficient vector of the polynomial, and the degree of freedom of the filter can be adjusted by setting the order K. The larger K is, the higher the order of the frequency response function that can be fitted, but this will increase the risk of model overfitting.

For computing the eigenvalues of the Laplacian matrix $L$, matrix decomposition is typically necessary, which entails complex calculations. However, the Chebyshev polynomial method, as indicated in [18], circumvents the need for decomposition of $L$. This method leverages a series of Chebyshev polynomials that can be computed iteratively, with the k-th term defined as:

$$T_k(x) = 2xT_{K-1}(x) - T_{K-2}(x), T_0(x) = 1, T_1(x) = x \quad (11)$$

Use truncated K-order Chebyshev polynomials to represent graph convolution (graph filtering) operations:

$$f_\theta(\Lambda) = \sum_{k=0}^{K} \theta_k T_k(\bar{\Lambda}), \bar{\Lambda} = \frac{2\Lambda}{\lambda_{\max}} - I \qquad (12)$$

The filtered Graph signal is:

$$Y = f_\theta(\bar{L})X = \sum_{k=0}^{K} \theta_k T_k(\bar{L})X, \bar{L} = \frac{2(I - L_{sym})}{\lambda_{\max}} - I \quad (13)$$

In order to reduce the receptive field of each layer and simplify the model, let the one-layer graph convolution layer $K = 1$ and $\theta = \theta_0 = \theta_1$, then the single-layer graph convolution can be simplified as:

$$Y = \theta(I + L_{sym})X = HX \qquad (14)$$

Since the range of eigenvalues for the symmetric normalized Laplacian matrix $L_{sym}$ spans $[-1, 1)$ [19], using it directly in the frequency response function $(1 + \lambda_{sym})^K$ t can cause the frequency band signal of $L_{sym} > 1$ to diverge, leading to potential issues like exploding or vanishing gradients. To mitigate signal component divergence from this frequency response, Kipf et al. [21] introduced a technique known as renormalization, which effectively adds a self-loo、p to each node. This can be represented as:

$$\begin{cases} \tilde{A} = A + I \\ \tilde{D} = D + I \\ \tilde{L}_{sym} = \tilde{D}^{-\frac{1}{2}} \tilde{A} \tilde{D}^{\frac{1}{2}} \end{cases} \qquad (15)$$

The final graph convolution can be expressed as:

$$Y = \sigma(\tilde{L}_{sym} X \Theta), \Theta \in R^{f \times f_{out}}, Y \in R^{N \times f_{out}} \qquad (16)$$

The normalized adjacency matrix $\tilde{L}_{sym}$, now including self-loops, is applied to the left of the graph signal vector $X$, and the trainable linear transformation $\Theta$ alters the number of channels or integrates information across each channel. To further bolster the network's expressive power, a nonlinear mapping $\sigma$ is employed. The frequency response function acts as a contraction map, functioning as a low-pass filter on the signal, which helps to dampen high-frequency noise and distill useful information within the low-frequency domain.

### D. Few-shot learning

In the current academic landscape, few-shot learning (FSL) is typified by data segmentation techniques that partition the original dataset into distinct meta-training $D_{train}$ and meta-testing $D_{test}$ sets, ensuring that $D_{train} \cap D_{test} = \phi$; This approach, which emphasizes task-oriented learning, centers on the concept of meta-tasks: independent tasks each designed to reflect an $N-way, K-shot$ learning scenario, simulating a small-sample environment. For each meta-task, N categories are sampled from the meta-training set, with K samples per category forming a support set $S_N$ and an additional set $Q_N$ for queries, containing Q samples per category. The support and query sets are meticulously structured to avoid overlap. The support set and query set can be expressed as $S_N = \{(x_i, y_i) | i = 1, 2, ..., N \times K\}$,

$Q_N = \{(x_i, y_i) | i = 1, 2, ..., N \times Q\}$, where $S_N \cap Q_N = \phi$ is satisfied. The model adapts its global parameters with each meta-task, aiming to achieve robust generalization across different meta-tasks and ultimately facilitating effective learning from small samples.

## III. METHOD

In this section, the multi-scale graph convolution filtering (MSGCF) in this article will be introduced. Its main flow chart is shown in **Figure 1**. This section introduces 3 main parts: data preprocessing, multi-scale graph convolution filtering, and training objectives.

### A. Data preprocessing

After collecting the original one-dimensional signal, because the data dimension is too high, in order to reduce the parameter complexity of the subsequent graph convolution network, data preprocessing needs to be performed first. 2D-CNN can effectively reduce the dimensionality and extract features of multi-channel two-dimensional arrays with huge amounts of data. After reconstructing the original signal into a two-dimensional array $x_0$, after $k$ layers of conv2d, the output is:

$$y_k = \text{Re}Lu(Maxpool(b_k + conv2D(W_k, y_{k-1}))) \quad (17)$$

Where $b_k$ and $W_k$ are the bias vector and convolution kernel parameters of the k-th layer, $y_k$ is the output of the k-th layer, satisfying $y_0 = x_0$, $Maxpool$ is the maximum pooling, and $\text{Re}Lu$ is the activation function (nonlinear mapping).

After completing feature extraction and dimensionality reduction, the data is divided into a small sample set - a support set and a query set, using the $N-way, K-shot$ method. The label features of the support set are label hot encoding of N categories, and the query set is all 0 encoding. The final features of the support set samples and query set samples are the concatenation of the final feature $x$ output by CNN2D and the label feature, expressed as:

$$x_{support, query} = x \oplus one - hot(label) \quad (18)$$

where $X_{support, query}$ is the final initial features of the support set and query set, which are used to build the initial graph.

### B. Multi-scale graph convolution filter

Since graph convolution behaves as low-pass filtering in the frequency domain and aggregates subgraph information in the spatial domain, increasing the number of graph convolution layers can aggregate more node information and more receptive fields in the spatial domain, which is beneficial to the utilization of limited resources. sample information, but increasing the number of layers and excessive low-pass filtering will make the information difference of nodes smaller, resulting in over-smoothing. Multi-scale graph convolution filtering (MSGCF) achieves a balance between improving the receptive field and over-smoothing by fusing the input and output of the previous layer when graph convolution layers are connected in series and globally paralleling single-layer graph convolution. The specific structure is as follows The MSGCF module is shown in **Figure 1**.

First, the initial graph after data preprocessing only has feature information and no structural information. In order to be able to perform graph convolution operations, a graph structure needs to be constructed. First define the weight matrix $W \in R^{N \times N \times f_{in}}$ of the graph to satisfy:

$$W_{ij} = |X_{input}(i) - X_{input}(j)| \quad (19)$$

where $X_{input} \in R^{N \times f_{in}}$ is the input graph feature matrix, the weight matrix W describes the Manhattan distance between nodes, and $X_{input}$ is expressed as:

$$X_{input} = [x_{query1}, ..., x_{queryN \times Q}, x_{support1}, ..., x_{supportN \times K}] \quad (20)$$

Use conv2D to perform weight and channel transformation on W. In the same way as equation (17), the reconstructed Laplacian matrix $\tilde{L}_{sym} \in R^{N \times N}$ can be obtained. The output of one layer of GCN can be expressed as:

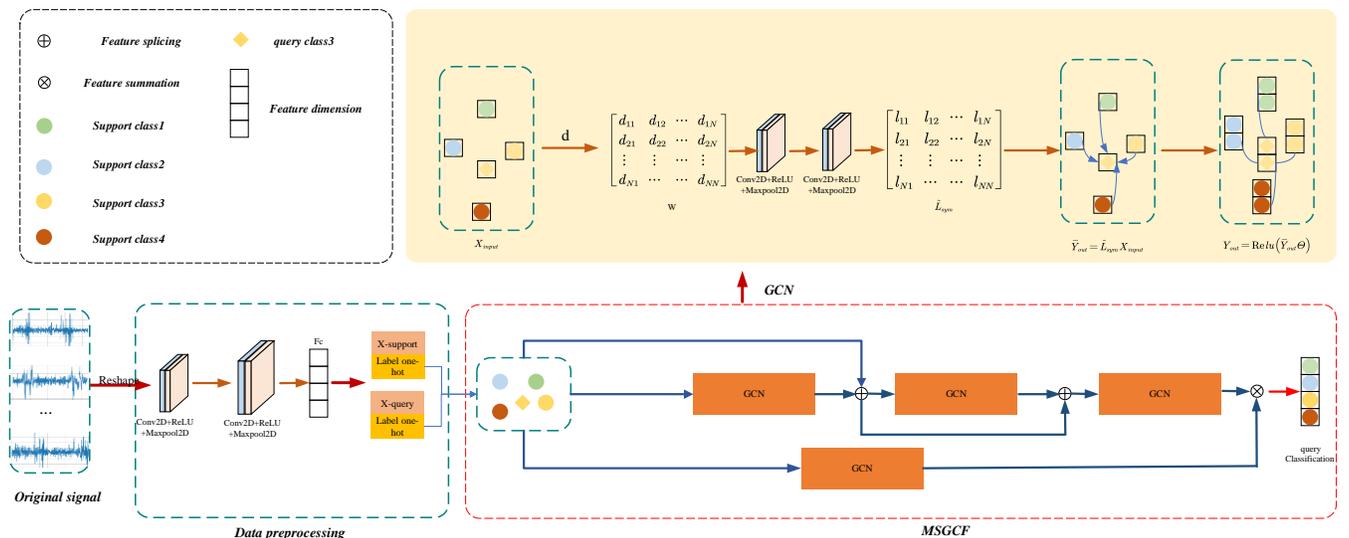

**Fig. 1.** Multi-scale graph convolution filtering structure

$$X_k = \text{Re}Lu\left(\tilde{L}_{sym-k} X_{k-1} \Theta_k\right) \quad (21)$$

where $X_k$ is the output of the k-th layer GCN, $\tilde{L}_{sym-k}$ is the $\tilde{L}_{sym}$ of input graph of the k-th layer, and $\Theta_k$ is the linear transformation matrix of the feature channels of the k-th layer, with learnable parameters.

MSGCF can be expressed as two parts, the series part is called the local channel, and the parallel part is called the global channel. The local channel improves the receptive field by stacking GCN, and can alleviate the over-smoothing phenomenon by splicing input features and output features, and aggregates more K-order subgraph information, where the local channel can be expressed as:

$$X_k = GCN(X_{k-1} \oplus X_{k-2}) \quad (22)$$

The global channel directly maps the query set category labels through parallel single-layer GCN. Since only first-order subgraphs are aggregated, most of the global information is retained. The parallel connection method avoids excessive low-pass filtering due to excessive local channel stacking. The convergence phenomenon of node information. From the perspective of deep learning, the local channel and the global channel are in a state of confrontation. Therefore, the optimal result is the best balance between improving the receptive field and over-smoothing the signal. The final label output can be expressed as:

$$\hat{y}_{query} = x_{K-query} \otimes GCN(X_0)[query] \quad (23)$$

*C. Training objectives*

In order to optimize MSGCF in this article and find suitable parameters, the loss function adopts the cross-entropy loss function. All parameters are trained end-to-end. The loss function is expressed as:

$$L = -\sum_{i=1}^{N} I(y_{query} == i) \log(\hat{y}_{query}[i]) \quad (24)$$

where $y_{query}$ is the true label of the query sample. When the query sample is the i-th category, $I(x)=1$ and the rest $I(x)=0$, $\hat{y}_{query}$ is the predicted label of the query sample output by MSGCF, and $\hat{y}_{query}[i]$ is the probability of belonging to the i-th category.

IV. EXPERIMENT

In this section, the MSGCF method proposed in this paper is evaluated on the Paderborn University Bearing Dataset (PU). This section mainly describes three main parts: introduction to data sets and experimental settings, comparison with other methods, and ablation experimental analysis. All experiments in this article were completed using Python on a Windows PC with R5-5600 4.7GHz and RTX 3080 (12GB).

*A. Introduction to data sets and experimental settings*

This study employs the Paderborn University (PU) dataset to mirror industrial scenarios characterized by data scarcity, variable operating conditions, and complex fault modes. The dataset encompasses 4 distinct operating conditions and 13 types of compound faults, with a high sampling frequency of 64 kHz. Operating conditions are simulated by varying the rotational speed of the driving system, applying radial force to bearings, and altering the load torque. The dataset delineates four specific operating conditions, with each of the 13 fault types detailed by damage mode (specifically fatigue), bearing component involvement, combination of damages, extent of damage, and damage traits. For each fault type, spread across four working conditions, there are 20 samples, aligning with industrial scenarios of limited data. For comprehensive details, readers are directed to the PU dataset documentation [21].

To assess the precision of the MSGCF method, this paper categorizes the original dataset into 52 (13x4) classes, with each class containing 20 samples. The dataset is split in an 8:2 ratio, assigning 41 classes to the training set and 11 to the test set. Ablation studies utilize a 5-way, 5-shot approach, whereas comparison experiments are conducted with 5-way, 1-shot and 5-way, 5-shot configurations for segregating support and query sets.

*B. Ablation experimental analysis*

To evaluate the efficacy of MSGCF, an ablation study was initially conducted by excluding the global module and solely employing the local module to ascertain the optimal number of layers for peak performance. Theoretically, the performance is expected to improve with an increasing number of layers before reaching a peak and subsequently declining. The optimal layer count determined through experimentation is then utilized in tandem with the global module to finalize the MSGCF framework discussed in this paper. As depicted in **Figure 2**, the accuracy on the test set varies with the layer count. The peak accuracy of 83.11% is achieved with three layers. With just two layers, the model's receptive field is inadequate, leading to underutilization of the limited sample information and a slight decrease in accuracy by 2.38%. Beyond three layers, a decline in accuracy is observed, attributable to over-smoothing from excessive low-pass filtering—resulting in a 0.82% drop with four layers and a further 2.46% reduction with five layers.

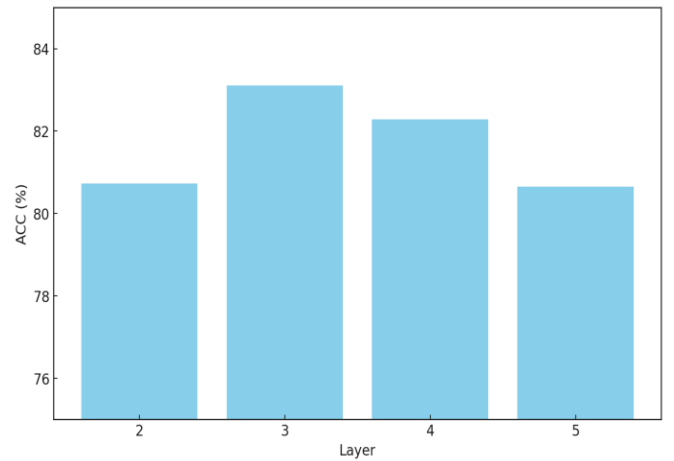

**Fig. 2.** Performance comparison of different layers

Consequently, within these experimental conditions, the MSGCF architecture is realized by integrating three serially connected layers of GCN for local channels and a single layer of GCN in parallel for global channels. Detailed comparisons from the ablation study are presented in **Table I**. Employing the control variable method for analysis reveals that both the local and global channels in MSGCF contribute to performance enhancement, registering a 3.19% improvement over the baseline GNN scenario.

TABLE I. ABLATION EXPERIMENT ON 5-WAY, 5-SHOT

| Name | Contains local channels | Contains global channels | Number of layers | ACC |
|---|---|---|---|---|
| GNN | ✗ | ✗ | 3 | 81.86% |
| GNN | ✓ | ✗ | 2 | 80.73% |
| GNN | ✓ | ✗ | 3 | 83.11% |
| GNN | ✓ | ✗ | 4 | 82.29% |
| GNN | ✓ | ✗ | 5 | 80.65% |
| **MSGCF** | ✓ | ✓ | 3 | **85.05%** |

*C. Comparison with other methods*

To further substantiate the superiority of the MSGCF method for few-shot diagnosis, this study benchmarks it against other methods—MAML, TBPN, and WDCNN—within the realm of few-shot learning. The comparative results, as illustrated in **Table II**, demonstrate that MSGCF outperforms in both Task 51 and Task 55. Unlike TBPN and WDCNN, which limit message propagation to support and query samples, MSGCF extends propagation within support samples, thereby enhancing small sample information utilization. Additionally, it effectively mitigates over-smoothing by utilizing local and global channels, resulting in heightened accuracy. This advantage underscores the potential of GNNs in few-shot learning contexts when over-smoothing is adequately controlled.

TABELE II. COMPARISON WITH OTHER METHODS

| Method | $5-way, 1-shot$ | $5-way, 5-shot$ |
|---|---|---|
| MAML | 77.35% | 80.76% |
| TBPN | 76.11% | 78.62% |
| WDCNN | 72.36% | 76.81% |
| **MSGCF** | **82.94%** | **85.05%** |

V. CONCLUSION

This study introduces the Multi-Scale Graph Convolution Filtering (MSGCF) method, a novel approach for few-shot diagnosis tailored for industrial applications. MSGCF adeptly balances enhancing the receptive field against the risk of over-smoothing inherent in stacked Graph Convolution Networks (GCNs). Its efficacy is demonstrated through experiments conducted on the Paderborn University (PU) dataset, where it showcased superior performance. This method offers valuable insights and potential research avenues for industrial fault diagnosis.

REFERENCES

[1] B. Yang, Y. Lei, F. Jia, et al., "An intelligent fault diagnosis approach based on transfer learning from laboratory bearings to locomotive bearings," Mechanical Systems and Signal Processing, vol. 122, pp. 692-706, 2019..

[2] H. Lv, J. Chen, T. Pan, et al., "Hybrid attribute conditional adversarial denoising autoencoder for zero-shot classification of mechanical intelligent fault diagnosis," Applied Soft Computing, vol. 95, 106577, 2020.

[3] J. Zhang et al., "A Novel Deep DPCA-SVM Method for Fault Detection in Industrial Processes," 2019 IEEE 58th Conference on Decision and Control (CDC), Nice, France, 2019, pp. 2916-2921.

[4] M. Wang, F. Cheng, K. Chen, J. Mi, Z. Xu and G. Qiu, "Incipient Fault Detection with Feature Ensemble Based on One-Class Machine Learning Methods," 2023 62nd IEEE Conference on Decision and Control (CDC), Singapore, Singapore, 2023, pp. 4867-4872.

[5] B. Song, S. Tan, H. Shi, et al., "Fault detection and diagnosis via standardized k nearest neighbor for multimode process," Journal of the Taiwan Institute of Chemical Engineers, vol. 106, pp. 1-8, 2020.

[6] K. Zhang, J. Chen, T. Zhang, et al., "A compact convolutional neural network augmented with multiscale feature extraction of acquired monitoring data for mechanical intelligent fault diagnosis," Journal of Manufacturing Systems, vol. 55, pp. 273-284, 2020.

[7] H. Zhao, S. Sun, B. Jin, "Sequential fault diagnosis based on LSTM neural network," IEEE Access, vol. 6, pp. 12929-12939, 2018.

[8] T. Pan, J. Chen, J. Xie, Y. Chang, Z. Zhou, "Intelligent fault identification for industrial automation system via multi-scale convolutional generative adversarial network with partially labeled samples," ISA Trans., vol. 101, pp. 379-389, 2020.

[9] S. Zhang, F. Ye, B. Wang and T. G. Habetler, "Few-Shot Bearing Anomaly Detection via Model-Agnostic Meta-Learning," 2020 23rd International Conference on Electrical Machines and Systems (ICEMS), Hamamatsu, Japan, 2020, pp. 1341-1346.

[10] C. Jiang, H. Chen, Q. Xu, et al., "Few-shot fault diagnosis of rotating machinery with two-branch prototypical networks," Journal of Intelligent Manufacturing, vol. 34, no. 4, pp. 1667-1681, 2023.

[11] D. Lee, J. Jeong, "Few-shot learning-based light-weight WDCNN model for bearing fault diagnosis in siamese network," Sensors, vol. 23, no. 14, 6587, 2023.

[12] V. Garcia, J. Bruna, "Few-shot learning with graph neural networks," arXiv preprint arXiv:1711.04043, 2017.

[13] N. Wang, M. Luo, K. Ding, et al., "Graph few-shot learning with attribute matching," in Proceedings of the 29th ACM International Conference on Information & Knowledge Management, 2020, pp. 1545-1554.

[14] L. Yang, L. Li, Z. Zhang, et al., "DPGN: Distribution propagation graph network for few-shot learning," in Proceedings of the IEEE/CVF Conference on Computer Vision and Pattern Recognition, 2020, pp. 13390-13399.

[15] X. Yu, B. Tang, and K. Zhang, "Fault diagnosis of wind turbine gearbox using a novel method of fast deep graph convolutional networks," IEEE Transactions on Instrumentation and Measurement, vol. 70, pp. 1-14, 2021.

[16] C. Yang, J. Liu, Q. Xu, et al., "A generalized graph contrastive learning framework for few-shot machine fault diagnosis," IEEE Transactions on Industrial Informatics, 2023.

[17] D. I. Shuman, S. K. Narang, P. Frossard, et al., "The emerging field of signal processing on graphs: Extending high-dimensional data analysis to networks and other irregular domains," IEEE Signal Process. Mag., vol. 30, no. 3, pp. 83-98, 2013.

[18] M. Defferrard, X. Bresson, and P. Vandergheynst, "Convolutional neural networks on graphs with fast localized spectral filtering," Advances in Neural Information Processing Systems, vol. 29, 2016.

[19] T. N. Kipf, M. Welling, "Semi-supervised classification with graph convolutional networks," arXiv preprint arXiv:1609.02907, 2016.

[20] C. Li, S. Li, H. Wang, et al., "Attention-based deep meta-transfer learning for few-shot fine-grained fault diagnosis," Knowledge-Based Systems, vol. 264, 110345, 2023.

[21] C. Lessmeier, J. K. Kimotho, D. Zimmer, et al., "Condition monitoring of bearing damage in electromechanical drive systems by using motor current signals of electric motors: A benchmark data set for data-driven classification," in PHM Society European Conference, vol. 3, no. 1, 2016.